\pdfoutput=1

\documentclass[11pt]{article}

\usepackage[]{EMNLP2023}
\usepackage{times}
\usepackage{latexsym}
\usepackage{graphicx}
\usepackage{xspace}
\usepackage{microtype}
\usepackage{multirow}
\usepackage{inconsolata}
\usepackage[normalem]{ulem}
\usepackage{amsmath}
\usepackage{titlesec}
\usepackage{array}
\usepackage{caption}

\newcommand{\td}{\texttt{text-davinci-003}}
\newcommand{\cd}{\texttt{code-davinci-002}}

\usepackage{times}
\usepackage{latexsym}
\usepackage{graphicx}
\usepackage{subcaption}
\usepackage[T1]{fontenc}
\usepackage{array}
\usepackage[utf8]{inputenc}
\usepackage{booktabs}
\usepackage{microtype}
\usepackage{multirow}
\usepackage{inconsolata}
\usepackage[normalem]{ulem}
\usepackage{amsmath}
\usepackage{titlesec}
\usepackage{caption}

\titlespacing{\paragraph}{%
  0pt}{
  0.2 \baselineskip}{
  1em}%

\usepackage[T1]{fontenc}

\usepackage[utf8]{inputenc}

\usepackage{microtype}

\usepackage{inconsolata}

\title{Evaluating Factual Consistency of Summaries with \\ 
Large Language Models}

\author{Shiqi Chen$^{*1,2}$\quad 
  Siyang Gao$^2$\quad
  Junxian He$^{\dagger1}$ \\
  $^1$Hong Kong University of Science and Technology \quad  $^2$City University of Hong Kong  \\
  \texttt{schen438-c@my.cityu.edu.hk, siyangao@cityu.edu.hk, junxianh@cse.ust.hk}
  }
\begin{document}
\maketitle

\renewcommand{\thefootnote}{\fnsymbol{footnote}}
\footnotetext[1]{Work done during Shiqi's visit to HKUST.}
\footnotetext[2]{Corresponding author.}
\renewcommand{\thefootnote}{\arabic{footnote}}

\begin{abstract}

Detecting factual errors in summaries has been an important and challenging subject in summarization research.
Inspired by the emergent ability of large language models (LLMs), we explore evaluating factual consistency of summaries by directly prompting LLMs. 
We present a comprehensive empirical study to assess the ability of LLMs as factual consistency evaluators, which consists of (1) analyzing different LLMs such as the GPT model series and Flan-T5; (2) investigating a variety of prompting methods including vanilla prompting, chain-of-thought prompting, and a sentence-by-sentence prompting method to tackle long summaries; and (3) evaluating on diverse summaries generated by multiple summarization systems, ranging from pre-transformer methods to SOTA pretrained models.
Our experiments demonstrate that prompting LLMs is able to outperform the previous best factuality systems in all settings, by up to 12.2 absolute points in terms of the binary classification accuracy on inconsistency detection.\footnote{Code is available at \url{https://github.com/hkust-nlp/llmeval_sum_factual}.}

\end{abstract}

\section{Introduction}
While recent progress of conditional generation has improved the text summarization performance dramatically~\citep{lewis-etal-2020-bart,zhang2020pegasus,liu-etal-2022-brio}, the factuality problem -- where models often yield summaries that are not grounded by the source input -- remains prominent and critical in abstractive summarization systems. 
For example, prior research found that 30\% of automatic summaries could contain hallucinations~\citep{goodrich2019assessing,kryscinski-etal-2019-neural}, and this phenomenon persists even in the state-of-the-art pretraining-based models~\citep{cao-wang-2021-cliff}. 
Unfortunately, such factuality errors cannot be reflected by the traditional summarization metrics like ROUGE scores~\citep{lin-2004-rouge}. Thus, a reliable and effective factual consistency\footnote{We use ``factual consistency'', ``faithfulness'', and ``factuality'' exchangeably throughout the paper.} evaluation method is desired. 
\begin{figure}[!t]
    \centering
    \includegraphics[width=\columnwidth]{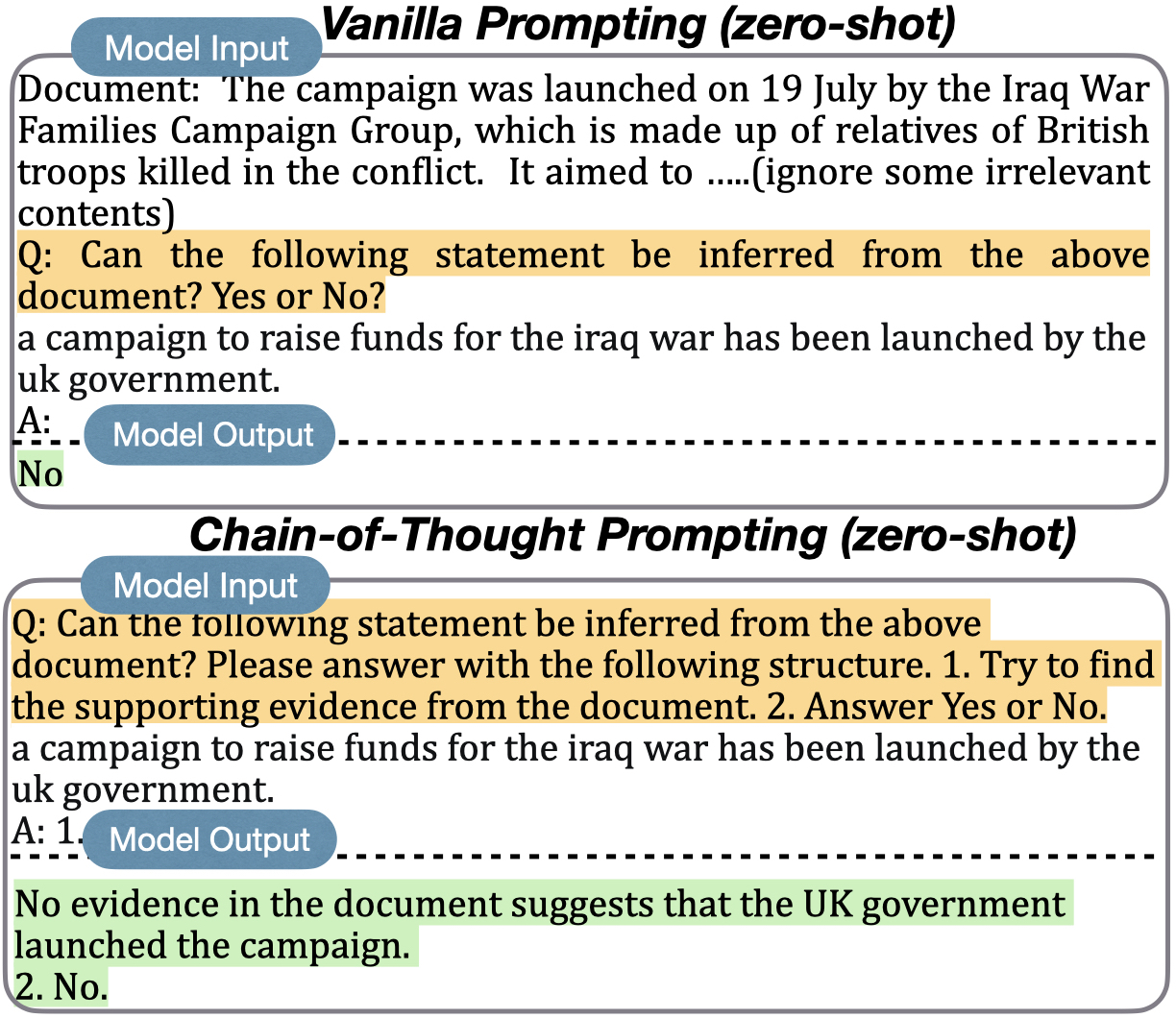}
    \caption{An example of prompting the \texttt{text-davinci-003} language model for factuality evaluation. Chain-of-thought prompting adds more instructions to ask the model to find the evidence.}
    \label{fig:intro}
    \vspace{-10pt}
\end{figure}

Different from previous work that focuses on training specific natural language inferences or question answering models for factuality evaluation~\citep{goodrich2019assessing,wang2020asking,kryscinski2020evaluating,durmus2020feqa,scialom2021questeval}, 
we explore an alternate approach through directly prompting LLMs. This work is inspired by the recent success of zero/few-shot prompting with LLMs~\citep{liu2021pre}, particularly on instructing tuning~\citep{sanh2022multitask,wei2022finetuned,ouyang2022training,chung2022scaling} 
and chain-of-thought prompting~\citep{wei2022chain} 
which greatly boost the prompt understanding and reasoning abilities of LLMs.
Given these latest advances, in this paper, we aim to answer: are LLMs off-the-shelf factuality evaluators for summarization?

\begin{table*}[!t]
    \centering
    \vspace{-3pt}
    \resizebox{1.0 \textwidth}{!}{
     \scriptsize
    \begin{tabular}{ccc}
    \toprule
       {\bf Source Document} & {\bf Prompt} & {\bf Model Output}\\
       \midrule
       \multirow{3}{*}{\parbox{0.25\textwidth}{\vspace{-20pt}
       
       (CNN)The classic video game "Space Invaders" was developed in Japan back in the late 1970's -- and now their real-life counterparts are the topic of an earnest political discussion in Japan's corridors of power. Luckily, Japanese can sleep soundly in their beds tonight as the government's top military official earnestly revealed that the country's Air Self Defense Force (ASDF) had never encountered an extraterrestrial unidentified flying object....(ignore some contents)
       \textbf{Inoki has appeared in the U.S.-based WWE -- which describes him as "among the most respected men in sports-entertainment" -- and is the founder of the New Japan Pro Wrestling organization.}He entered Japan's Upper House for a second stint in politics in 2013.....(ignore some contents)}} & \parbox{0.5\textwidth}{\emph{Vanilla Prompting:}
       
       Q: Can the following statement be inferred from the above document? Yes or No?

`` space invaders '' is the founder of the new japan pro wrestling organization . inoki has appeared in the u.s.-based wwe -- which describes him as `` among the most respected men in sports-entertainment '' .

A:} & \parbox{0.2\textwidth}{No.}  \\
      \cmidrule{2-3}
        & \parbox{0.5\textwidth}{\emph{Chain of thought prompting:}
        
        Q: Can the following statement be inferred from the above document? Please answer with the following structure. 1. Try to find the supporting evidence from the document. 2. Answer Yes or No.

`` space invaders '' is the founder of the new japan pro wrestling organization . inoki has appeared in the u.s.-based wwe -- which describes him as `` among the most respected men in sports-entertainment '' .

A: 1.}& \parbox{0.2\textwidth}{Antonio Inoki is described as "among the most respected men in sports-entertainment" and is the founder of the New Japan Pro Wrestling organization. 
2. No, Space Invaders is not the founder of the New Japan Pro Wrestling organization.} \\
        \cmidrule{2-3}
        &\parbox{0.5\textwidth}{\emph{Sentence-by-sentence prompting:}
        
        Q: Can the following statement be inferred from the above document? Yes or No?

1. `` space invaders '' is the founder of the new japan pro wrestling organization .

2. inoki has appeared in the u.s.-based wwe -- which describes him as `` among the most respected men in sports-entertainment '' .

A: 1.} & \parbox{0.2\textwidth}{No. 2. Yes.} \\
        
     \bottomrule
    \end{tabular}}
    \caption{A zero-shot prompting example with three prompting methods. We bold the content from the source that supports the prediction. The listed prompts do not show the prepended documents for ease of space. We use the exact wording of instructions here in our experiments.}
    \label{tab:prompt}
    \vspace{-3pt}
\end{table*}

To this end, we cast factual consistency evaluation as an entailment task and directly query LLMs whether the summary can be inferred by the source document or not as shown in Figure~\ref{fig:intro}. 
We incorporate recent developments of LLMs and prompting techniques:
(1) Beyond the vanilla prompting, 
we examine chain-of-thought prompting, which encourages LLMs to articulate a reasoning process for their predictions. Furthermore, we introduce a sentence-by-sentence prompting method designed for long summaries, breaking down lengthy generations into multiple abbreviated summaries, thereby simplifying the task.
(2) We consider five powerful LLMs with different sizes and accessibility: GPT-4~\citep{openai2023gpt4}, ChatGPT~\citep{chatgpt}, \texttt{text-davinci-003}, \texttt{code-davinci-002}, and the open-source Flan-T5~\citep{chung2022scaling} model that has 11 billion parameters and can be deployed with reasonable hardware requirements.

Concurrent research investigates the use of LLMs to evaluate generated text. While~\citet{fu2023gptscore} emphasize general text generation tasks and different aspects, we focus specifically on the factuality of summarization. Furthermore, their approach necessitates access to the logits of generated sequences, which are often unavailable in current LLMs' access.~\citet{luo2023chatgpt} assess ChatGPT as the factuality evaluator in a manner similar to our method. In comparison, however, our study spans a broader range of LLMs that outperform ChatGPT significantly and introduces a novel sentence-by-sentence prompting approach that yields considerable improvements in most settings. ~\citet{gekhman2023trueteacher} have proposed a novel approach to generate synthetic data using LLMs. By evaluating the performance of LLMs fine-tuned on the generated data, the authors successfully demonstrated the efficiency of both the data and the model. This study highlights the significant contribution of LLMs in advancing the field of summary generation and evaluation.

In the experiments, we evaluate factuality evaluators on five factuality benchmarks that cover different quality levels of summaries, which are generated from a variety of summarization systems.
Empirical results show that prompting LLMs outperforms the existing factuality evaluation methods in all settings. The improvement over best of the compared methods is up to 12.2 absolute points in terms of binary inconsistency classification accuracy.
Our results imply that LLMs are better factual consistency evaluators of summaries, which provides supportive evidence for a potential shift in the methodology of factuality evaluation to prompting LLMs.
\section{Prompting Methods and Models}
\subsection{Prompting}
Following previous work, we cast factual consistency evaluation as an entailment task ~\citep{kryscinski2020evaluating, goyal2020evaluating, zhao2020reducing}, and ask the model whether the summary could be inferred from the source document.\footnote{We note that the formulation as an entailment task is not necessary -- we also tried asking whether the summary is factually correct or not directly, which works reasonably well as shown in Appendix~\ref{appdix:prompt}.}
Denote the document as $x$, the summary as $y$, and the human-written instruction as $i$. Then the prompt $p \equiv \langle x, i , y\rangle$, where $\langle\rangle$ represents concatenation of the document, instruction, and summary. 
We feed $p$ as input to the model that is expected to produce short, yes or no answers. We term such a vanilla version of prompting as \emph{vanilla prompting}. In the following, we describe another two more advanced prompting techniques, while an example of all the prompt formats we studied is illustrated in Table~\ref{tab:prompt}:

\begin{itemize}
 \item {\bf Chain-of-thought prompting:}~\citet{wei2022chain} demonstrate that it is helpful to ask the model to generate a thought process along with the final prediction. 
 Therefore, we design a chain-of-thought instruction, which aims to guide the model to output the supporting evidence from the document before making the final judgment.
 \item {\bf Sentence-by-sentence prompting:} Summaries often consist of multiple sentences and facts that need to be verified.  
 We design a framework to decompose the summary into smaller text blocks first and then evaluate them one by one.
 For simplicity, we just decompose the summary by sentence boundary that is already effective empirically,
 while we leave the study of other decomposition methods (e.g.~decomposition through prompting LLMs) as future work.
\end{itemize}

Table~\ref{tab:prompt} only exemplifies prompts under zero-shot prompting setting, while we also experiment with few-shot prompting for the three prompting methods. 
The few-shot demo examples are randomly picked from the validation set.
We have included more details on our prompt engineering practice in Appendix~\ref{appdix:prompt}.
\subsection{Models}
We study five LLMs: GPT-4~\citep{openai2023gpt4}, ChatGPT~\citep{chatgpt}, \texttt{text-davinci-003}, \texttt{code-davinci-002},  and Flan-T5~\citep{chung2022scaling}. Details are described in Appendix~\ref{appdix:models}.

\section{Experiments}
\begin{table*}[!t]
    \small
        \centering
        \vspace{-3pt}
        \newcolumntype{L}[1]{>{\raggedleft\arraybackslash}p{#1}}
        \newcolumntype{R}[1]{>{\raggedright\arraybackslash}p{#1}}
        \newcolumntype{P}[1]{>{\centering\arraybackslash}p{#1}}
        \resizebox{1 \textwidth}{!}{
        \begin{tabular}
    {@{}R{0.2\linewidth}R{0.1\linewidth}L{0.19\linewidth}L{0.19\linewidth}L{0.2\linewidth}@{}}
        
        \toprule
        \multirow{3}{*}{\textbf{Models}}                           & \multirow{3}{*}{\textbf{Prompt}} & \multicolumn{3}{c}{\bf{Dataset}}           \\ 
         \cmidrule(r){3-5}

                                          &         & SummEval    & XsumFaith   & XsumSota   \\ 
        \midrule
        \multicolumn{5}{c}{\emph{Previous Approaches}} \vspace{3pt} \\
        DAE                               &       --    & 69.9         & --   &     72.8         \\
        QuestEval                         &     --      & 71.3            &  59.7 &   66.6   \\
        SummaC-ZS                         &     --   & 67.7      & 52.1  &    56.5   \\
        SummaC-Conv                       &      --      & 73.7          &   66.0 &   63.1      \\
        QAFactEval                        &      --      & 76.6         &  60.2 &      66.0      \\
        \midrule
        \midrule
        \multirow{3}{*}{Flan-T5}          & vanilla    & {\bf85.2} / {\bf 78.7} &  58.6 / 58.0 &   {\bf 75.1} / {\bf 74.7} \\
                                          & cot       & 67.7 / 52.6 & 55.0 / 57.6 &   61.5 / 60.3 \\
                                          & sbs       & 70.9 / 75.3 &  --           &      --       \\
        \midrule
        \multirow{3}{*}{\texttt{code-davinci-002}} & vanilla    & 61.3 / 76.4 &  53.2 / 65.2 &   53.5 / 59.8\\
                                          & cot       & 56.1 / 72.1  & 52.3 / {\bf 68.8} &   51.6 / 54.0\\
                                          & sbs       & {\bf 76.6} / {\bf 86.3} &  --           &     --        \\
        \midrule
        \multirow{3}{*}{\texttt{text-davinci-003}} & vanilla    & {\bf 81.5} / {\bf 84.6} & 60.3 / 65.2  & {\bf 74.1} / 67.2 \\
                                          & cot       & 62.2 / 72.6 &  {\bf 66.8} / {\bf 69.0}  & 65.5 / 59.2  \\
                                          & sbs       & {\bf 83.4} / {\bf 88.0} &  --           &   --         
        \\
        \midrule
        \multirow{3}{*}{ChatGPT} & vanilla    &  65.3 / 68.9& {\bf 67.5} / {\bf 67.2}   & 63.3 / 65.2  \\
                                          & cot       &  59.9 / 68.5  & \underline{{\bf 69.7}} / 66.0   &  70.1 / 67.0   \\
                                          & sbs       & {\bf 83.3} / {\bf 80.0}  &  --           &   --   \\
        \midrule
        \multirow{1}{*}{GPT-4$^{\dagger}$} & --   & \underline{{\bf 88.8}}  &  {\bf 67.2}  &  \underline{{\bf 75.2}} \\
        \bottomrule
        \end{tabular}}
        \caption{
            Balanced accuracy (\%) on the test split. The results of LLMs are in the format of zero-shot/few-shot. Cot denotes chain-of-thought prompting and sbs denotes sentence-by-sentence prompting. Sbs prompting is applied to SummEval only since the summary in XSum contains just one sentence. We bold the numbers that exceed all the previous approaches, and underline the best accuracies on each dataset. We exclude DAE on XSumFaith for a fair comparison since it is trained on the human-annotated data from XSumFaith. 
            GPT-4$^{\dagger}$ is assessed with zero-shot vanilla prompting on XSum datasets and 2-shot sbs promtping on CNNDM datasets, due to cost consideration.
            Numbers of previous approaches are from~\citet{tang2022understanding}.
            } 
            \label{tab:cnndm_xsum_eval}
        \end{table*}
\vspace{8pt}
\subsection{Benchmark Datasets}
\label{sec:dataset}
A summarization faithfulness benchmark is composed of source documents, model-generated summaries, and annotated faithfulness labels. 
The formal faithfulness benchmarks mainly use two popular summarization datasets, CNN/Dailymail (CNNDM,~\citet{hermann2015teaching}) and XSum~\citep{nallapati2016abstractive}. 
CNNDM is a multi-sentence summarization dataset for CNN and Dailymail articles. Its reference summaries highly overlap with the source articles, resulting in a low degree of abstractiveness~\citep{zhang-etal-2018-abstractiveness}.
In contrast, summaries in XSum are typically more abstractive,
consequently leading summarization models to be more susceptible to generating factual errors within XSum~\citep{cao-wang-2021-cliff}.
Due to the disparate characteristics of CNNDM and XSum, we assess factual evaluators on them separately. 

Recently,~\citet{tang2022understanding} aggregates existing faithfulness benchmarks to form a new benchmark, AggreFact. 
We manually investigate the benchmarks included in AggreFact, and select a subset of them as described below as our testbed which are either commonly used or annotated by the authors or experts:\footnote{We found that some datasets annotated only by crowd-sourced annotators could be less reliable and contain annotation errors. The difficulties of crowd-sourcing high-quality annotation in the text summarization domain were also observed in~\citet{fabbri-etal-2021-summeval} by showing a large difference between crowd-sourced and expert annotations.}
SummEval~\citep{fabbri-etal-2021-summeval}, XsumFaith~\citep{maynez2020faithfulness}, Goyal21~\citep{goyal2021annotating}, CLIFF~\citep{cao-wang-2021-cliff}, FactCC~\citep{kryscinski2020evaluating}, and Frank~\citep{pagnoni2021understanding}.
These benchmarks cover summaries generated from a wide range of models ranging from pre-transformer methods to SOTA pretrained models.
In this paper, we combine Goyal21 and CLIFF -- where the summaries are produced by BART~\citep{lewis-etal-2020-bart} or T5~\citep{raffel2020exploring} -- as a benchmark to study the ability of evaluators to assess high-quality summaries produced by SOTA models, we refer to this combined benchmark as \emph{XSumSota}. 
 We distinguish XSumSota and XSumfaith to echo the findings in~\citet{tang2022understanding} that the performance of faithfulness evaluation methods degrades dramatically as the summaries are from more effective models. 
Details of these benchmarks are described in Appendix~\ref{appdix:bench}.


\subsection{Setup}
\label{sec:metric}
We apply greedy decoding to obtain output from LLMs in all settings unless otherwise specified. 
We run sentence-by-sentence prompting in CNNDM datasets only since the summary of the XSum dataset is a single sentence. 
For ease of spaces, we move the results on FactCC and Frank to Appendix~\ref{appdix:other_res}. In few-shot settings, within a benchmark we prepend the same two randomly picked demo examples (one is a positive example and the other is negative) from the validation set to the original prompt.\footnote{In our task, two example sequences are already over 1000 tokens due to the long source document.}
We perform analysis on the effect of the number of demo examples as well as error types in Appendix~\ref{appdix:analysis}.

\paragraph{Metric:}
We use balanced accuracy~\citep{brodersen2010balanced} as the evaluation metric following previous work~\citep{laban-etal-2022-summac,tang2022understanding}, which is defined as:
\begin{equation}
    BAcc = \frac{1}{2}\Big(\frac{TP}{TP+FN}+\frac{TN}{TN+FP}\Big),
\end{equation}
where TP stands for True Positive, FN is False Negative, TN is True Negative, and FP is False Positive.
Random predictions would obtain a 50\% balanced accuracy score.
Different from most prior approaches which need to tune a threshold hyperparameter to convert raw output scores into binary labels~\citep{tang2022understanding}, the LLMs directly produce discrete, yes or no predictions as shown in Table~\ref{tab:prompt}.
\subsection{Baselines}
We compare LLMs with five top-performing evaluators: DAE~\citep{goyal2020evaluating}, QuestEval~\citep{scialom2021questeval}, QAFactEval~\citep{fabbri2021qafacteval}, SummC-ZS~\citep{laban-etal-2022-summac} and SummaC-Conv~\citep{laban-etal-2022-summac}. Detailed description can be found at Appendix~\ref{appdix:baselines}.

\subsection{Results}
\paragraph{Are LLMs better factual consistency evaluators?}
The full results comparing different evaluators across the three benchmarks are illustrated in Table~\ref{tab:cnndm_xsum_eval}.
We see that LLMs achieved the state-of-the-art performance on all benchmarks. The improvements over the previous best on SummEval, XsumFaith, XsumSota are 12.2, 3.7, and 2.4 absolute points respectively. 
\texttt{text-davinci-003} and GPT-4 are the most effective models overall, outperforming the non-LLM approaches on all the three benchmarks.
Flan-T5, \texttt{code-davinci-002} and ChatGPT beat the previous best on two out of three benchmarks. 
Therefore, we conclude that LLMs are indeed better factual consistency evaluators when properly prompted. 

\paragraph{Comparing different prompting methods:}
As shown in Table~\ref{tab:cnndm_xsum_eval}, chain-of-thought (cot) prompting hurts performance dramatically compared to vanilla prompting in most cases.
This is probably because the factual consistency task is less reasoning-intensive compared to numerical and symbolic reasoning tasks where cot archives success. 
Sentence-by-sentence (sbs) prompting clearly improves over vanilla prompting on SummEval for \cd~and~\td, particularly in~\cd, sbs is around or over 10 points better than vanilla prompting in both zero- and few-shot settings.
This verifies that decomposing a long summary into smaller blocks makes factual consistency evaluation easier. 

\paragraph{Comparing few-shot with zero-shot:}
While few-shot prompting fails to yield consistent gains over zero-shot prompting on all settings, 
it especially helps \texttt{code-davinci-002}, for example, it outperforms zero-shot prompting by $\sim$15/16/10 points on SummEval with vanilla/cot/sbs prompting, by 6.3 points on XSumSota with vanilla prompting, and by $\sim$12/16 points on XSumFaith with vanilla/cot prompting. 
After examining the model output, we found that \texttt{code-davinci-002} often fails to understand and follow the instructions in the zero-shot setting, but is able to do so when provided with exemplars.

\paragraph{Comparing different LLMs:}
Based on the previous comparisons, we summarize that \texttt{text-davinci-003} and GPT-4 are two best models for the factual consistency task, while being less sensitive to the availability of exemplars.
On the other hand, \texttt{code-davinci-002} requires providing a few demo examples to potentially work well. 
Importantly, Flan-T5 achieves surprising results in general -- under a zero-shot setting in SummEval and XSumSota, Flan-T5 not only beats all the baselines, but also outperforms both the GPT-3.5 variants that are orders of magnitude larger. 
The unsatisfying performance of Flan-T5 on XSumFaith may be due to a lack of per-dataset prompt tuning for XSumFaith. 


\paragraph{Are we there yet?}
While LLMs make great progress in factual consistency evaluation as shown in Table~\ref{tab:cnndm_xsum_eval}, we observe distinct patterns.
Taking \texttt{text-davinci-003} as an example, it has pushed the ACC of the CNNDM benchmark SummEval to 88\%, but its performance on the two XSum benchmarks is no more than 75\%. 
These results imply that it remains challenging to evaluate the faithfulness of highly abstractive summaries.
Therefore, faithfulness evaluation has to continue relying on human labor in practice at the current stage, and automatic metrics still have a long way to go.

\label{sec:bibtex}

\section{Discussion}
In this paper, we focus on one of the central tasks in summarization, factual consistency evaluation of summaries, and explore to prompt large language models to address it.
We perform a comprehensive empirical study demonstrating large language models are better factual consistency evaluators when properly prompted. 
We note that prompting LLMs is a highly flexible approach and could go beyond the usage in this paper for factuality consistency evaluation. 
\bibliography{anthology,custom}
\bibliographystyle{acl_natbib}
\clearpage
\newpage
\appendix



\appendix




\section{Analysis on Different Prompts}
\label{appdix:prompt}

\begin{table}[!h]
 \centering
\resizebox{0.9 \columnwidth}{!}{
\begin{tabular}{c}
\toprule
\midrule
\parbox{0.45\textwidth}{       
        Q: Can the following statement be inferred from the above document? Yes or No?
} \\
\midrule
\parbox{0.45\textwidth}{       
        Q: Is the following statement factually consistent with the above document? Yes or No?
} \\
\midrule
\parbox{0.45\textwidth}{       
        Q: Does the above document entail the following statement? Yes or No?
} \\
\midrule
\bottomrule
\end{tabular}}
\caption{The three different instructions we use to conduct the robustness experiment.}
\label{fig:appdix:prompt}
\end{table}

\paragraph{Prompt Engineering: }
LLMs are notoriously known to be sensitive to the precise wording of prompts, and thus prompt engineering is required in the prompting process. 
We try several instructions in our experiments and select the best one in terms of the validation performance, while we perform robustness analysis for different prompts. 
Note that we deliberately avoid the use of the term ``summary'' in the instruction but replace it with ``statement'', it is because we found that the term ``summary'' would reveal that the generated text is intended to be a summary of the source, and consequently, the model is inclined to function as a general summarization evaluation task rather than focusing only on factual consistency. 
We emphasize that we use the same instructions across all models and benchmarks without tuning them separately for each dataset.

\begin{table}[!t]
    \centering
    \resizebox{1 \columnwidth}{!}{
    \begin{tabular}{lcrrr}
    \toprule
    
     \multirow{2}{*}{\textbf{Model}}&\multirow{2}{*}{\textbf{Setting}}&\multicolumn{3}{c}{\bf{Dataset}}           \\ 
    \cmidrule(r){3-5}&
          & SumE& XSF& XSS \\
     \midrule
     \multirow{2}{*}{Flan-T5}& 0-shot& 76.7{$\pm$}5.0& 60.1{$\pm$}2.4 &   72.2{$\pm$}4.2 \\
     &  2-shot& 79.5{$\pm$}4.2  &  58.5{$\pm$}1.0  & 73.4{$\pm$}1.1 \\
     \midrule
       \multirow{2}{*}{\texttt{code}}& 0-shot& 65.0{$\pm$}10.4& 51.8{$\pm$}2.0& 47.7{$\pm$}6.3 \\
        &   2-shot&78.9{$\pm$}6.7  &  59.8{$\pm$}5.3  & 58.5{$\pm$}2.3 \\
         \midrule
         \multirow{2}{*}{\texttt{text}}& 0-shot&   84.9{$\pm$}2.0  &  60.5{$\pm$}0.5  & 73.9{$\pm$}0.6 \\
         &   2-shot&87.8{$\pm$}1.0 &  65.2{$\pm$}0.3  & 66.6{$\pm$}1.4 \\
    \bottomrule
    \end{tabular}}
    \caption{
        Mean and standard deviation of balanced accuracy (\%) on SummEval (SumE), XSumFaith (XSF), and XSumSota (XSS). We use sentence-by-sentence prompting for SumE and vanilla prompting for XSF and XSS. \texttt{code} and \texttt{text} are short for \texttt{code-davinci-002} and \texttt{text-davinci-003} respectively. 
        }
    \label{tab:robust}
    \vspace{-14pt}
    \end{table}

\paragraph{Robustness on different prompts:}
We run experiments using the three prompts listed in Table~\ref{fig:appdix:prompt}, and report the mean and the standard deviation of balanced accuracy.
We evaluate Flan-T5, \texttt{text-davinci-003}, and \texttt{code-davinci-002} on SummEval, XSumFaith, and XSumSota.
For few-shot settings, we also randomly shuffle the order of the exemplars in addition to varying the instruction. 
We utilize vanilla prompting on XSumFaith and XSumSota, and sentence-by-sentence prompting on SummEval. 
Results are reported in Table~\ref{tab:robust}.
\texttt{text-davinci-003} exhibits the smallest variance, demonstrating strong abilities to follow different, synonymous instructions. 
\texttt{code-davinci-002} and Flan-T5 are more sensitive to the wording of prompts.
\section{Models}
\label{appdix:models}
Below we describe the 5 LLMs that we study.
\paragraph{GPT-4}
~\citep{openai2023gpt4} is the newest and most powerful model in GPT-family. The model weights are not released and we use it through the OpenAI API.
\paragraph{ChatGPT}
~\citep{chatgpt} is a sibling model for instructGPT. It's trained on plenty of instructions and responses by supervised learning and it then went through Reinforcement Learning from Human Feedback (RLHF). These two techniques enable it to follow human instructions efficiently.
We also examine ChatGPT through API.
\paragraph{\texttt{text-davinci-003}}
is a variant in the GPT-3.5 series. It is obtained after joint pretraining of text and code and then further tuned with annotated instruction data\footnote{\href{https://beta.openai.com/docs/model-index-for-researchers}{https://beta.openai.com/docs/model-index-for-researchers}} -- consequently, \texttt{text-davinci-003} is much more powerful than the original GPT-3 model~\citep{brown2020language}. 
\texttt{text-davinci-003} is likely to have 175 billion parameters following GPT-3 but cannot be verified from public information.
We also use it through API.
\paragraph{\texttt{code-davinci-002}}
is another GPT-3.5 variant. While it was intended to be used in the code domain, recent work indicates that \texttt{code-davinci-002} is more effective than the text davinci models on text numerical reasoning tasks~\citep{zhou2022least}. 
Similar to \texttt{text-davinci-003}, we examine \texttt{code-davinci-002} through API.

\paragraph{Flan-T5}
\citep{chung2022scaling} is fine-tuned from the T5 model~\citep{raffel2020exploring} on around 2000 NLP datasets with instruction tuning, demonstrating great performance on a variety of tasks through prompting. 
We experiment with the largest Flan-T5 model, \texttt{flan-t5-xxl}.
Notably, \texttt{flan-t5-xxl} has released model weights and is 11-billion-parameter large, a much smaller size compared to GPT-3.5.
Flan-T5 is probably the most capable language model with prompting that is open-source and could be deployed in relatively common hardware conditions (e.g. two 40GB GPUs). 
We select Flan-T5 in our study to indicate what open-source, easy-to-deploy LLMs can achieve in factual consistency evaluation.
We use the transformers package~\citep{wolf-etal-2020-transformers} to evaluate Flan-T5.

\section{Benchmark Datasets}
\label{appdix:bench}
\begin{table}[ht]
    \small
    \centering
    \resizebox{1 \columnwidth}{!}{
    \begin{tabular}{@{}p{0.28\linewidth}p{0.28\linewidth}p{0.14\linewidth}p{0.14\linewidth}@{}}
    \toprule
    \textbf{Dataset} & \textbf{Annotators} &\textbf{Val/Test}&\textbf{Pos/Neg} \\
    \midrule
    SummEval \newline \cite{fabbri-etal-2021-summeval} & 5 crowd-sourced annotators and 3 authors &800/798 & 719/79\\
    \midrule
    XSumFaith \newline \cite{maynez2020faithfulness} & 3 trained annotators  &1000/853 & 60/793\\
    \midrule
    XsumSota \newline \cite{cao-wang-2021-cliff,goyal2021annotating} & 2 experts or 2 authors &200/200 & 89/111 \\
    \midrule
    FactCC \newline \cite{kryscinski2020evaluating} &  2 authors &931/503 & 441/62 \\
    \midrule
    Frank \newline \cite{pagnoni2021understanding} &  3 crowd-sourced annotators &671/1575 & 529/1046 \\
    \bottomrule
    \end{tabular}}
    \caption{Metadata of the three benchmarks that we focus on. XSumSota is a combined benchmark of~\citet{cao-wang-2021-cliff} and \citet{goyal2021annotating} for summaries generated by the state-of-the-art summarization models.}
    \label{tab:metadata}
    \end{table}



\paragraph{SummEval~\citep{fabbri-etal-2021-summeval}}
is the most complete faithfulness benchmark for CNNDM as far as we know. The summaries are from 16 different models including both pre-transformer models and the state-of-the-art summarization models such as BART~\citep{lewis-etal-2020-bart}, PEGASUS~\citep{zhang2020pegasus}, and T5~\citep{raffel2020exploring}.
The annotations are from 5 crowd-sourced annotators and 3 authors of the benchmark.

\paragraph{XSumFaith~\citep{maynez-etal-2020-faithfulness}}
is the most commonly used faithfulness benchmark for XSum~\citep{fabbri2021qafacteval,laban-etal-2022-summac,zhou2021detecting}.
It contains summaries generated from 5 models which do not include the SOTA summarization models. 
The transformer-based models studied in XSumFaith are GPT-2~\citep{radford2019language} and BERT~\citep{devlin-etal-2019-bert}. 
The annotations are from 3 trained annotators. 

\paragraph{Goyal21~\citep{goyal2021annotating}}
contains both the CNNDM and the XSum samples. The two authors of this work manually annotated the summaries. 
We use the XSum split in this dataset where all the annotated summaries are generated from a tuned BART model. 

\paragraph{CLIFF~\citep{cao-wang-2021-cliff}}
consists of summaries generated by SOTA models (including T5 and BART). The annotations are from 2 experts. Similar to Goyal21, we take the XSum split from this dataset.

\paragraph{FactCC~\citep{kryscinski2020evaluating}}
consists of summaries generated by pre-transformer models. Two authors of this work annotated this dataset. It contains only CNNDM samples.

\paragraph{Frank~\citep{pagnoni2021understanding}}
consists of summaries generated by various summarization models from pre-transformer models and SOTA models. Three crowd-sourced annotators  annotated this dataset. It contains both CNNDM and Xsum samples.

We have noticed that there is another benchmark SummaC~\citep{laban-etal-2022-summac}, which is an integration of six datasets including CoGenSumm~\citep{falke2019ranking}, XsumFaith, Polytope~\citep{huang2020have}, FactCC, SummEval, and Frank. Here we do not include SummaC as a whole since the CoGenSumm benchmark ranks pairs of generated summaries rather than detecting factually consistent summaries, and pairs of summaries can be both factually consistent or inconsistent. Also, some annotated error types in Polytope such as addition, omission, or duplication are unrelated to factuality errors by definition. 
As a result, we think that the SummaC benchmark as a whole may not be suitable for factuality evaluation, as mentioned in~\citet{tang2022understanding} as well.  Note that we do not separate the SOTA summaries out in SummEval since there are only 3 negative samples out of 200 SOTA test samples in total -- SOTA models rarely make factual errors on less abstractive summarization, and we think it is not representative either to use just 3 negative samples to characterize the inconsistency detection ability of evaluators. 
The benchmark metadata is shown in Table~\ref{tab:metadata}.


\section{Analysis}
\label{appdix:analysis}

\begin{figure}[!h]
    \centering
    \begin{subfigure}[b]{1\columnwidth}
        \includegraphics[width=1\textwidth]{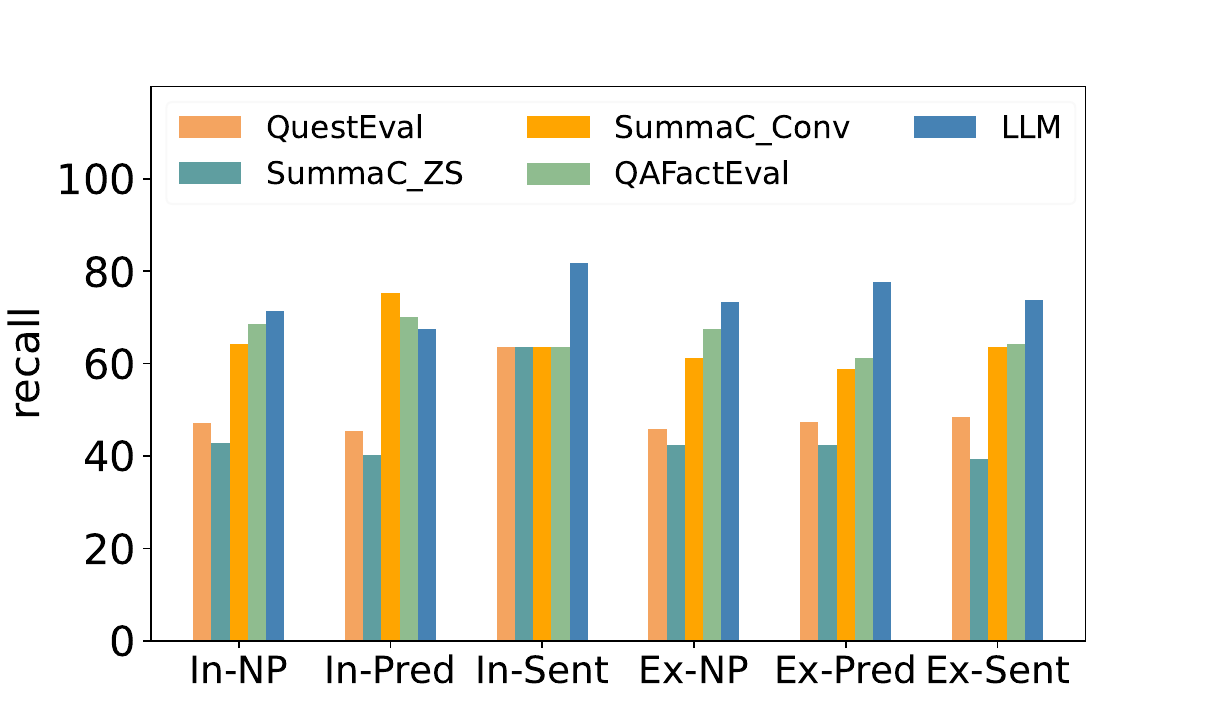}
    \end{subfigure}
    \begin{subfigure}[b]{1\columnwidth}
        \vspace{1pt}
        \includegraphics[width=1\textwidth]{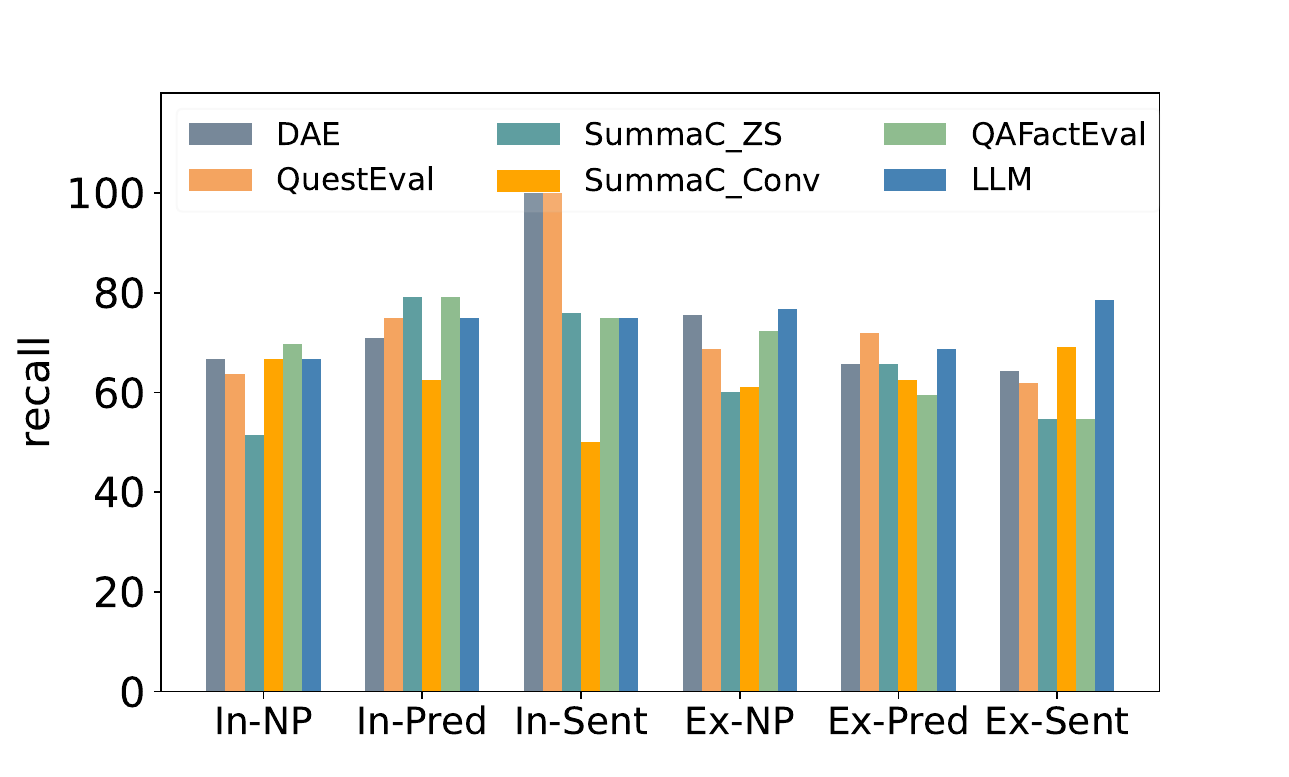}
    \end{subfigure}
            
    \caption{Recall of the identified different types of factual errors from the evaluators on XSumFaith (top) and XSumSota (bottom). We exclude DAE on XSumFaith for a fair comparison since it is trained with the human annotations from XSumFaith.}
    \label{fig:error-type}
    \vspace{-14pt}
    \end{figure}

\paragraph{Fine-grained analysis on different types of factuality errors:}


After an overview of the balanced accuracy results, we perform a more fine-grained analysis on the different types of factuality errors detected to obtain a deeper understanding of the evaluators' predictions. 
We resort to the error type annotations from AggreFact that aggregates error type definitions from prior work and establishes a unified factuality error type scheme~\citep{tang2022understanding}. 
Specifically, it defines six factuality error types as a set $\{intrinsic, extrinsic\} \times \{noun phrase, predicate, sent\}$.
Intrinsic errors denote hallucinated content using the information in the source document, while extrinsic errors are synthesized generations that ignore the source document altogether. For example, introducing new nouns or verbs not related to the source text. 
$\{noun phrase, predicate, sent\}$ indicates the errors happen at a noun phrase, a predicate, or the entire summary. 
We refer the readers to~\citet{tang2022understanding} for more detailed explanations and examples of these error types. 
We report recall of the identified errors on XSumFaith and XSumSota,\footnote{There is no error type annotation for SummEval.} and compare \texttt{text-davinci-003} with the best prompting method (in terms of the overall balanced accuracy) on each dataset against the baselines. 
As shown in Figure~\ref{fig:error-type}, \texttt{text-davinci-003} identifies more errors in XSumFaith than all the baselines on 5 out of 6 error types. 
The results on XSumSota are more mixed, where \texttt{text-davinci-003} outperforms the baselines on 3 out of 6 error types.
These findings suggest a similar conclusion as in~\citet{tang2022understanding} that current factuality systems cannot be uniformly good at identifying every error type across datasets.
 
\paragraph{Effect of number of exemplars: }
\begin{figure}[!t]
\centering
    \begin{subfigure}[b]{0.49\columnwidth}
        \includegraphics[width=1\textwidth]{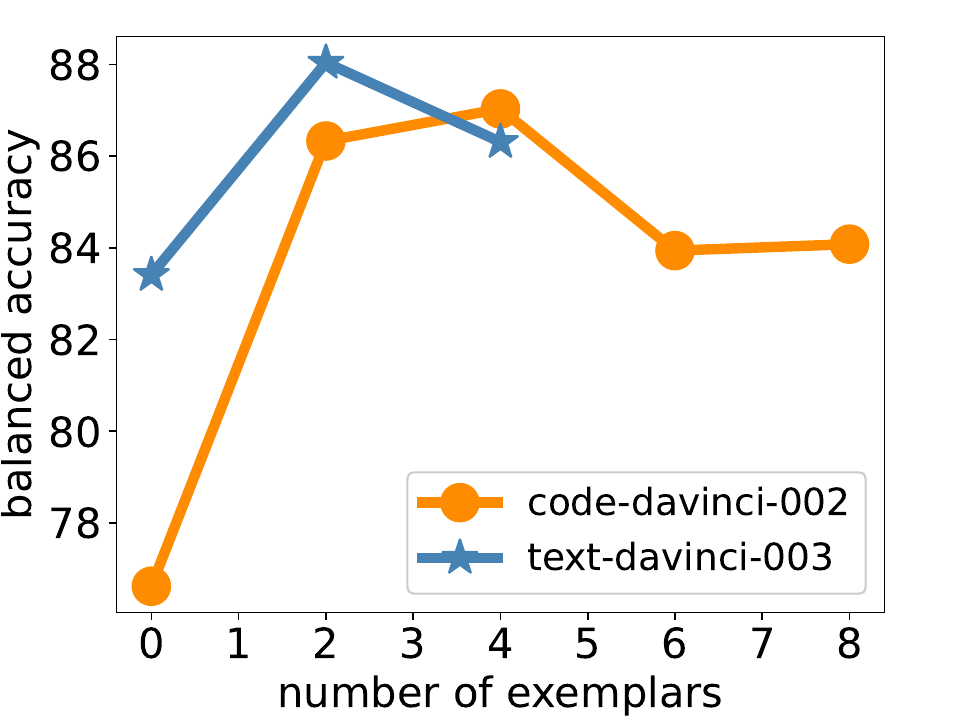}
    \end{subfigure}
    \hfill
    \begin{subfigure}[b]{0.49\columnwidth}
        \includegraphics[width=1\textwidth]{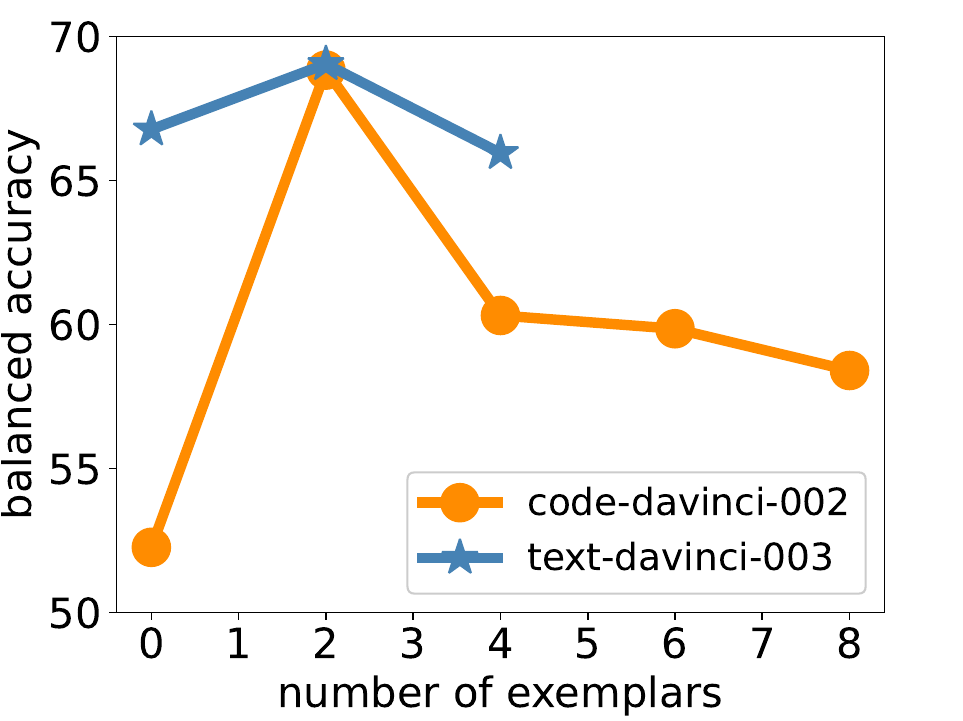}
    \end{subfigure}
        

\caption{
    Balanced accuracy (\%) on SummEval (left) and XSumFaith (right) varying the number of exemplars. We use up to 4 exemplars for \texttt{text-davinci-003} due to the context window size limit (4000 tokens).
    }
\label{fig:few-shot}
\end{figure}
We vary the number of exemplars on SummEval and XsumFaith where few-shot learning helps the most. 
We study \texttt{text-davinci-003} and \texttt{code-davinci-002}. 
Flan-T5 is excluded since more than 2 exemplars do not fit within its context window.
We adopt the best prompting method on each benchmark for the analysis -- sentence-by-sentence prompting and chain-of-thought prompting for SummEval and XSumFaith respectively. 
Results are shown in Figure~\ref{fig:few-shot}. 
The balanced accuracy is not monotonically increasing as we increase the number of shots. 
While the best performance of \texttt{code-davinci-002} is achieved with 4 shots on SummEval, 2 shots is the best configuration in other settings. 
This may be due to the long context of few-shot prompts in summarization.

\section{Our Baselines}
\label{appdix:baselines}
\paragraph{DAE~\citep{goyal2020evaluating}} 
is an arc-grained entailment-based evaluation method. It evaluates the factuality of each dependency arc in the generated summary separately and combines them as the final result. 

\paragraph{QuestEval~\citep{scialom2021questeval}}
is a QA-based approach which aggregates the answer overlap scores from questions generated from the summary and answered with the document, and from questions generated from the document and answered with the summary.

\paragraph{QAFactEval~\citep{fabbri2021qafacteval}} 
is another QA-based approach that computes the answer overlap scores from questions generated from the summary and answered with the document, but with improved components at each stage.

\paragraph{SummaC-ZS~\citep{laban-etal-2022-summac}}
is an entailement-based method which computes the maximum entailment score for each summary sentence, and aggregates all the scores through an averaging operation to obtain the final score.  
\paragraph{SummaC-Conv~\citep{laban-etal-2022-summac}}  
is an extension of SummaC-ZS where for each summary sentence, SummaC-Conv computes the entailment scores with respect to all the source sentences, passes the obtained scores as features to a convolution layer to produce the summary sentence score, and then averages as in SummaC-ZS.

We emphasize that we evaluate all the baselines in a threshold-per-dataset setting -- the baselines use a different threshold hyperparameter (detailed in~\textsection\ref{sec:metric} on each benchmark tuned separately, while the LLMs use the same instructions across all datasets.

\section{Results on another two benchmarks}
\label{appdix:other_res}
We report results of three most powerful LLMs on another two benchmarks: FactCC and Frank. 
We utilize sentence-by-sentence prompting for the CNNDM samples and vanilla prompting for XSum samples in both benchmarks.
The results are shown in Table~\ref{tab:2other bench}. Here the numbers of previous approaches for FactCC are from~\citet{tang2022understanding}, the numbers for Frank are from~\citet{laban-etal-2022-summac}. \
We can see the best performance is also achieved by LLMs on both two benchmarks.
\begin{table}[!h]
    \small
        \centering
        \newcolumntype{L}[1]{>{\raggedleft\arraybackslash}p{#1}}
        \newcolumntype{R}[1]{>{\raggedright\arraybackslash}p{#1}}
        \newcolumntype{P}[1]{>{\centering\arraybackslash}p{#1}}
        \resizebox{1 \columnwidth}{!}{
        \begin{tabular}
    {@{}R{0.3\linewidth}R{0.1\linewidth}L{0.2\linewidth}L{0.2\linewidth}@{}}
        
        \toprule
        \multirow{3}{*}{\textbf{Models}}                           & \multirow{3}{*}{\textbf{}} & \multicolumn{2}{c}{\bf{Dataset}}           \\ 
         
         \cmidrule(r){3-4}

                                          &    & FactCC & Frank \\ 
        \midrule
        \multicolumn{4}{c}{\emph{Previous Approaches}} \vspace{3pt} \\
        DAE                               &            &     70.4 &     61.7      \\
        QuestEval                         &            & 65.5& 82.1  \\
        SummaC-ZS                         &         &83.5&72.1  \\
        SummaC-Conv                       &             & -- &74.4     \\
        QAFactEval                        &             &      74.2&--     \\
        \midrule
        \midrule
       
        {\texttt{text-davinci-003}}                    
                                          &        &{\underline{\bf 84.9}} / 79.4&{\bf 84.5} / {\bf 83.4}  \\
        \multirow{1}{*}{ChatGPT} 
                                          &         & 71.9 / 77.9 & 81.6 / {\bf 83.2}  \\
        \multirow{1}{*}{GPT-4} &    &79.6& \underline{{\bf 87.9}}
        \\                             
        \bottomrule
        \end{tabular}}
\caption{
            Balanced accuracy (\%) on the test split of FactCC and Frank. The results of LLMs are in the format of zero-shot/few-shot. All LLMs are assessed by sentence-by-sentence prompting method. For GPT-4, we only run the 2-shot setting to save cost. We bold the numbers that exceed all the previous approaches, and underline the best accuracies on each dataset. We exclude SummaC-Conv on FactCC for a fair comparison since it has been trained with a synthetic dataset from FactCC, thus its performance on FactCC may not be directly comparable to others.
            }
            \label{tab:2other bench}
            \vspace{-15pt}
        \end{table}



\section{Related Work}
\paragraph{Factual consistency evaluation:}
Prior factuality evaluation approaches can be divided into entailment-based methods and question answering (QA) methods. Entailment-based methods aim to determine whether a summary is entailed by the original document or not. 
They often apply 
both semantically-variant and semantically-invariant transformations to the summaries to construct a classification dataset to train the evaluation model~\citep{goodrich2019assessing, kryscinski2020evaluating, goyal2020evaluating, zhao2020reducing}. 
 Relying on heuristic transformations cannot cover all types of factual errors in summarization, limiting its performance as mentioned in~\citet{kryscinski2020evaluating}.
On the other hand, QA methods automatically yield questions to probe the facts in the document or summary, and then assess whether the facts in the document and the summary are consistent by answering these questions~\citep{wang2020asking,durmus2020feqa,scialom2021questeval} 
These approaches need 
to train additional models 
for question generation, question answering, or answer comparison,
where corresponding annotations are required and the errors of intermediate components could propagate to the final predictions.  



\paragraph{Prompt-based learning:}
Prompts in the context of language models refer to text instructions concatenated with the test input (zero-shot) or with few exemplars and the test input (few-shot). 
Large language models are able to perform various tasks without tuning~\citep{radford2019language,brown2020language,liu2021pre} when the prompts are fed into as the input to signal the task. 
Instruction tuning futher lifts the ability of LLMs to follow instructions in the prompts~\citep{wei2022finetuned,sanh2022multitask,ouyang2022training,chung2022scaling,iyer2022opt}. 
Recently, chain-of-thought prompting is proposed to trigger the reasoning abilities of LLMs, by asking the model to explain the thinking process during generation~\citep{kojima2022large,wei2022chain}.
In addition to text prompts,~\citet{gao2022pal} and~\citet{chen2022program} introduce program-of-thought prompting to generate executable code to perform numerical reasoning tasks. 
\end{document}